# Bangla Text Dataset and Exploratory Analysis for Online Harassment Detection

Md Faisal Ahmed, Zalish Mahmud, Zarin Tasnim Biash, Ahmed Ann Noor Ryen, Arman Hossain, Faisal Bin Ashraf

Department of Computer Science and Engineering, Brac University, Bangladesh

*Abstract*— Being the seventh most spoken language in the world, the use of the Bangla language online has increased in recent times. Hence, it has become very important to analyze Bangla text data to maintain a safe and harassment-free online place. The data that has been made accessible in this article has been gathered and marked from the comments of people in public posts by celebrities, government officials, athletes on Facebook. The total amount of collected comments is 44001. The dataset is compiled with the aim of developing the ability of machines to differentiate whether a comment is a bully expression or not with the help of Natural Language Processing and to what extent it is improper if it is an inappropriate comment. The comments are labeled with different categories of harassment. Exploratory analysis from different perspectives is also included in this paper to have a detailed overview. Due to the scarcity of data collection of categorized Bengali language comments, this dataset can have a significant role for research in detecting bully words, identifying inappropriate comments, detecting different categories of Bengali bullies, etc. The dataset is publicly available at https://data.mendeley.com/datasets/9xjx8twk8p.

## I. INTRODUCTION

Cyberbullying or online harassment is the use of electronic correspondence such as the online networking platform to menace an individual, ordinarily by sending messages of an intimidating or compromising nature. Due to the spread use of internet and social networking sites, cyberbullying has become a major concern. It has been a very challenging research domain to understand the huge amount of text data, i.e., posts, comments, messages in social sites, so that preventive measures can be taken beforehand. Natural Language Processing is a branch of artificial intelligence that deals with the interaction between computers and humans utilizing the natural language . A definitive goal of NLP is to read, decode, comprehend, and make sense of the human dialects in a way that is significant. Bangla is the 7th most speaking language in the world. Huge amount of Bengali speakers use Bangla Language in online platform. So, it has been a dire need to analyze these Bangla texts from the social sites. Nevertheless, there is scarcity of Bangla text data with proper label. Therefore, in this work we have collected a vast amount of Bangla comments of different kinds of online harassment.

This dataset provides social media harassment comments in Bengali language, which is the seventh most spoken language in the world with more than 265 million speakers [1]. Therefore, in terms of analyzing online harassment this dataset will be immensely helpful. This dataset contains various types of bullying including the victim's gender and profession, which aids the researchers to conduct in depth analysis of relationship between online harassment and other metadata. These data also includes the number of likes or reacts to each of the online comment, which broadens the area to identify the trend that shows the social acceptance or normalization of appreciative as well as harassing comments. These data will be helpful to detect cyberbullying in Bengali language as there are about 30 million daily active Facebook users in Bangladesh and a huge proportion of this population often has to deal with online harassment and bullying [2]. These dataset can also be used for identifying bully words, detecting different categories of harassment, and several other works in this related field. The dataset can be used to train and evaluate computational models and techniques for automation to reduce cyber harassment.

## II. EXPERIMENTAL DESIGN

The data was obtained from the comment sections of public posts of some popular Facebook pages and anonymized in compliance with the Facebook Platform Policy for Developers [3]. We scrapped the comments from specific posts. Then the comments in Bengali language were filtered out from other English or mixed comments. The comments were also checked for duplicates. After getting rid of the duplicates, the comments were labelled into five categories where the bully category has four subcategories. If the comment was found to be harassing, it was then checked in which subcategory it logically belongs. The subcategories for bullies are sexual, threat, troll, and religious. If the comments did not fall under any criteria of harassment, it was labelled as not-bully. All the members repeated the process multiple times and the consensus decision was taken to ensure that the labels were correct.

## III. DATA DESCRIPTION

The dataset [4] attached with this article in an excel file, contains comments from the interaction section under public posts by celebrities, politicians, sportsmen on the Facebook platform. The total amount of comments collected is 44001. For the convenience of describing, Table 1 below refers to each column, which includes the variable name, variable type, description of the variable and the source or engineering behind it.

Table 1 describes variables of the dataset that is available in the given link and the data type of each variable. Variable types in Table 1 are categorical, numeric and text-based data.

Table 1: Variables of Dataset

| Variable | Type | Description |
|---|---|---|
| **Comment** | Text | Collected comment of the users |

| Category | Categorical | Occupation of the celebrities. For example, actor, singer, politician etc. |
|---|---|---|
| **Gender** | Categorical | Gender of the celebrities |
| **Number of Reacts** | Integer | Number of likes, reactions on that comment |
| **Label** | Categorical | Type of bully / non-bully |

## IV. EXPLORATORY ANALYSIS

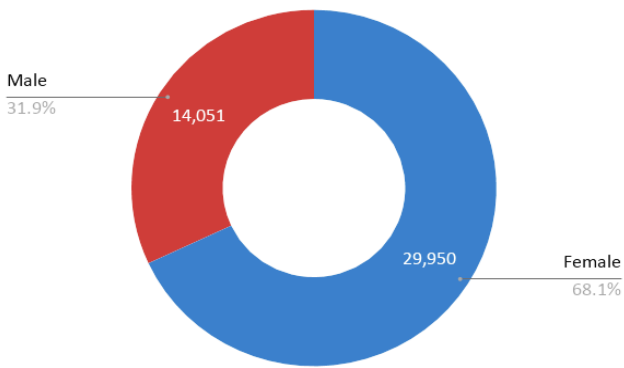

Fig. 1 Male to Female percentage ratio of the total number of data collected

Fig. 1 shows us the numbers of male and female victims based on the total number of data collected. According to our dataset, 31.9% i.e. 14,051 comments of the total comments, are targeted towards male victims and 68.1% i.e. 29,950 comments are aimed towards female victims (Table 2).

Table 2: Number of Male and Female victim

|  | Quantity | Percentage |
|---|---|---|
| **Male** | 14051 | 31.9% |
| **Female** | 29950 | 68.1% |

Fig. 2 conveys the percentage of comments depending on the profession of the victim. 21.31% i.e. 9,375 comments are targeted towards victims who are social influencers, 5.98% i.e. 2633 comments are targeted towards politicians, 4.68% i.e. 2061 comments are targeted towards athletes, 6.78% i.e. 2981 comments are targeted towards singers and 61.25% i.e. 26951 comments are targeted towards actors.

Table 3 conveys the summarization of the percentage of comments depending on the profession of the victim. The 'Quantity' column states the number of comments put by users and the 'Percentage' column states the respective quantity of the comments in percentage.

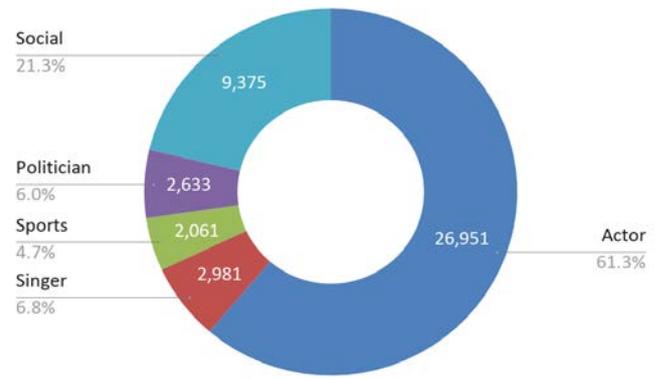

Fig. 2 Percentage of data based on the victim's profession

Table 3. Number of comments in different profession

|  | Quantity | Percentage |
|---|---|---|
| **Actor** | 26951 | 61.25% |
| **Social** | 9375 | 21.31% |
| **Politician** | 2633 | 5.98% |
| **Sports** | 2061 | 4.68% |
| **Singer** | 2981 | 6.78% |

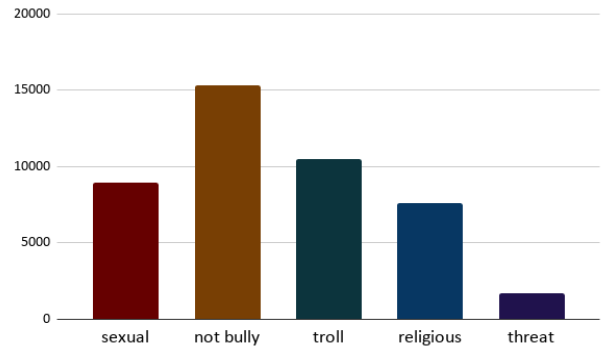

Fig. 3 Total number of comments based on the label

Fig. 3 shows the bar charts of the number of comments based on the label manually assorted to identify whether a comment is 'sexual', 'not bully', 'troll', 'religious', or 'threat'.

Table 4. Number of comments in different category of harassment

|  | Quantity | Percentage |
|---|---|---|
| **Sexual** | 8928 | 20.29% |
| **Not bully** | 15340 | 34.86% |
| **Troll** | 10462 | 23.78% |
| **Religious** | 7577 | 17.22% |
| **Threat** | 1694 | 03.85% |

Table 4 conveys the summarization of the total number of comments depending on the label. The 'Quantity' column states the number of comments put by users and the 'Percentage' column states the respective quantity of the comments in percentage.

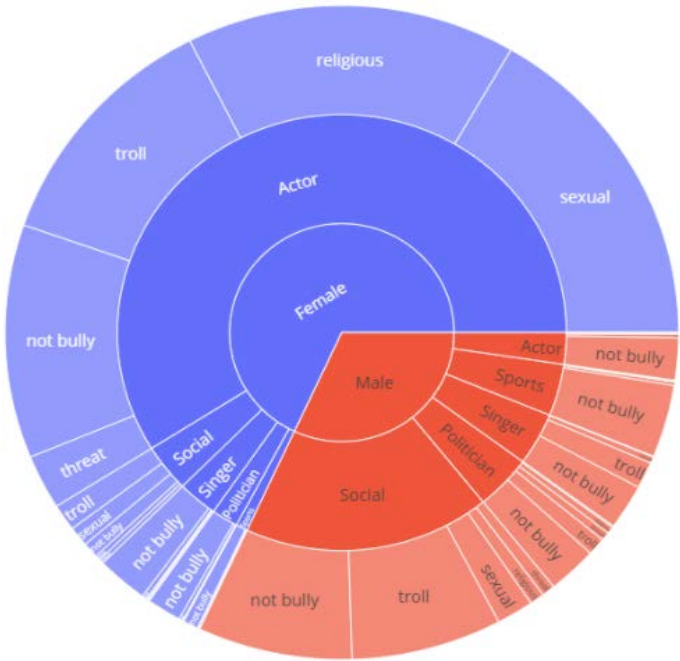

Fig. 4 Sunburst chart of hierarchy of data summary

Fig. 4 summarizes the whole dataset. It visualizes the number of comments according to the victim's gender, profession, and category of the comment. The root of the implemented hierarchy is the gender of the victim. The victim's gender is further analyzed according to their respective profession. Each of the professions are analyzed according to the category of the comments on the victim's post.

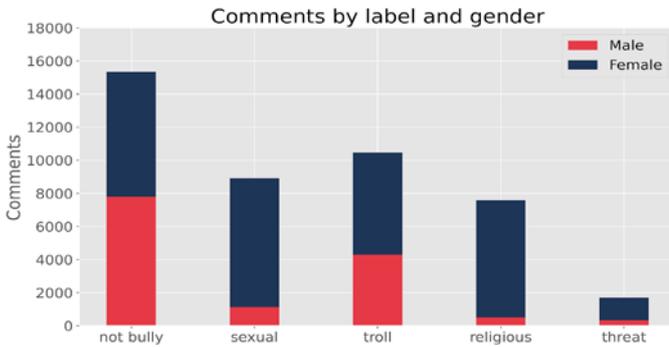

Fig. 5 Comments by label and gender

Fig. 5 shows us the comments by label and gender. The X-axis shows the number of comments; Y-axis shows the bar of comments under each label i.e. 'not bully', 'sexual', 'troll', 'religious' and 'threat' input by male users and female users.

Table 5 conveys the summarization of the total number of comments depending on the label and gender. The 'Male' column states the number of comments put by users on male victims and the 'Female' column states the number of comments put by users on female victims.

Table 5. Distribution of different harassment among victim's gender

|  | Male | Female |
|---|---|---|
| **Sexual** | 1132 (8.06%) | 7796 (26.03%) |
| **Not bully** | 7793 (55.46%) | 7547 (25.20%) |
| **Troll** | 4292 (30.55%) | 6170 (20.60%) |
| **Religious** | 491 (3.49%) | 7086 (23.66%) |
| **Threat** | 343 (2.44%) | 1351 (4.51%) |

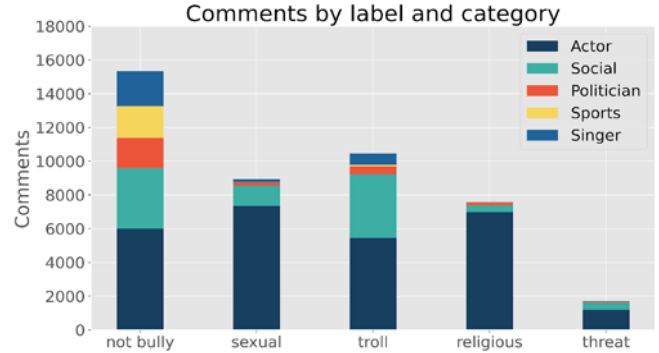

Fig. 6 Distribution of comments by label and category

Fig. 6 shows us the comments by label and category. The X-axis shows the number of comments; Y-axis shows the number of comments under each label i.e. 'not bully', 'sexual', 'troll', 'religious' and 'threat' from each categorical profession of the people i.e. 'Actor', 'Social', 'Politician', 'Sports', 'Singer' from which the data has been collected.

## V. CONCLUSION

This dataset is built with proper care and the data are fully anonymized. These data can help the researchers largely to identify online harassment in Bangla Text. The distribution and number of data in each category provide a good source of learning a good machine learning model to detect different kind of online harassment in Bangla language.